\documentclass[journal]{IEEEtran}
\usepackage{ amssymb  }
\usepackage{ mathdots }
\usepackage{ amsmath  }
\usepackage{ color    }
\usepackage{ graphicx }
\usepackage{ amsfonts }
\usepackage{ amssymb  }
\usepackage{ epstopdf }
\usepackage{ upgreek  }
\usepackage{ bm       } 
\usepackage{ graphicx }
\usepackage{ subfigure}
\usepackage{ mathtools}   
\usepackage{booktabs}
\usepackage{multirow}
\usepackage[final]{pdfpages}
\usepackage{bm}
\usepackage{cite}

\usepackage{array}
\usepackage{float}
\newcolumntype{L}[1]{>{\raggedright\let\newline\\\arraybackslash\hspace{0pt}}m{#1}}
\newcolumntype{C}[1]{>{\centering\let\newline\\\arraybackslash\hspace{0pt}}m{#1}}
\newcolumntype{R}[1]{>{\raggedleft\let\newline\\\arraybackslash\hspace{0pt}}m{#1}}

\begin{document}
\title{Scalable Learning Paradigms for Data-Driven Wireless Communication}
\author{Yue Xu, Feng Yin, Wenjun Xu, Chia-Han~Lee, Jiaru Lin, and Shuguang Cui
\thanks{Yue Xu, Wenjun Xu, and Jiaru Lin are with Key Lab of Universal Wireless Communications, Ministry of Education, Beijing University of Posts and Telecommunications, Beijing, 100876 China (e-mail:xuy@bupt.edu.cn, wjxu@bupt.edu.cn, jrlin@bupt.edu.cn).}
\thanks{Chia-Han Lee is with National Chiao Tung University, Hsinchu, Taiwan (e-mail: chiahan@nctu.edu.tw).}
\thanks{Feng~Yin and Shuguang~Cui are with the Chinese University of Hong Kong, Shenzhen and Shenzhen Research Institute of Big Data, Shenzhen 518172, China (Email: yinfeng@cuhk.edu.cn, robert.cui@gmail.com).}}

\maketitle

\begin{abstract}
The marriage of wireless big data and machine learning techniques revolutionizes the wireless system by the data-driven philosophy.
However, the ever exploding data volume and model complexity will limit centralized solutions to learn and respond within a reasonable time. Therefore, scalability becomes a critical issue to be solved.
In this article, we aim to provide a systematic discussion on the building blocks of scalable data-driven wireless networks. On one hand, we discuss the forward-looking architecture and computing framework of scalable data-driven systems from a global perspective. 
On the other hand, we discuss the learning algorithms and model training strategies performed at each individual node from a local perspective. 
We also highlight several promising research directions in the context of scalable data-driven wireless communications to inspire future research.
\end{abstract}
\vspace{-1em}
	
\section{Introduction}
The next-generation wireless networks are migrating from traditional designs based on statistical modeling to the data-driven paradigms based on big data and machine learning. On one hand, the ever-expanding and context-rich wireless big data contain valuable information that can help customize the wireless system in almost all aspects: among others, architecture design, resource management, and task scheduling. On the other hand, machine learning, as one of the most powerful artificial intelligence tools, constitutes strong learn-from-data capabilities to discover useful patterns, e.g., human behaviors, from data and use them for accurate prediction and planning. Therefore, the data-driven wireless networks are anticipated to combine the strength of both big data and machine learning to develop better understanding on both the networks and the users in order to deliver personalized and adaptive service commitments to embrace a more intelligent future.

While the popularity of the term ``data-driven" has been recently fueled by the growth of big data and computing power, scalability becomes increasingly important due to manifold driving demands, including adaptiveness, low-latency, low-complexity and privacy-preserving.
Nowadays, state-of-the-art solutions are mostly developed based on the centralized design, by assuming that a single node has full access to the entire dataset and possesses a sufficient amount of storage and computing power for data processing and decision-making. However, more and more new breeds of intelligent devices and delay-sensitive applications require real-time response with high reliability, e.g., brake control for self-driving vehicles, collision avoidance for drones, and motion perceptions for augmented/virtual reality~(AR/VR). These new emerging applications have sparked a huge amount of interests in developing scalable paradigms to deliver lower latency and superior robustness than the traditional centralized counterparts.
Moreover, the ever growing network size (e.g., due to network densification and the explosion of IoT devices), model size (e.g., depth and width of deep learning models), and data volume altogether lead to optimization tasks with unprecedented complexity, which require computation that surpasses a single node's computing capability. This also necessitates the use of scalable models to decompose a large optimization problem into smaller pieces to be handled in a  distributed manner.  
Although information is inevitably demanded for precisely personalized wireless applications, privacy-sensitive data is preferable not to be logged into a centralized center purely for the purpose of model training. Such security and privacy concerns make scalable solution a natural choice for keeping raw data on local devices~(e.g., user devices and third-party edge devices) while only exchanging computed updates, e.g., gradients, for information sharing. 

Although the idea of developing scalable solutions has been widely recognized~\cite{8304392, 8664622, xu2019load, 8030322}, effective scalable methods are still in their infancy. Existing works mainly focus on developing dedicated scalable algorithms for specific wireless applications~\cite{8664622, xu2019load}. In contrast, developing an entirely scalable data-driven system needs to address a myriad of fundamental challenges, including architecture design and computing framework adaptation from a \textit{global} perspective, and learning algorithms and training strategies selection for individual devices from a \textit{local} perspective.
The overarching goal of this work is to provide a systematic discussion on the data-driven scalability. As such, we first draw a futuristic scalable data-driven wireless network architecture, which orchestrates the \textit{in-cloud intelligence} and the \textit{on-device intelligence} built upon the cloud-related and edge-related wireless infrastructures, respectively. Such an intelligence-everywhere blueprint envisions how to properly integrate scalable intelligence into the design and operation of next-generation wireless networks. 
Then, we present parallel and fully distributed scalable learning frameworks which specify how distributed machines can collaborate with each other in a joint learning process. 
Next, we discuss the learning algorithms performed at each individual node. In particular, we draw public attention from the artificial neural network (ANN) based algorithms to the Bayesian nonparametric learning algorithms and reinforcement learning (RL) algorithms, which we believe can better handle the ever increasing uncertainty in various aspects. 
Furthermore, we envision the emerging and promising trend of knowledge transfer, which encourages to properly embed the distilled knowledge from the existing models~(e.g., traditional statistical models and experienced learners) into the training process so as to improve the learning performance at each individual node. 
Finally, we provide concrete use cases to give a quantitative demonstration of how to apply the presented scalable learning techniques to specific wireless applications. 

\section{Scalable Architecture for Data-driven Wireless Communication}
\label{sec:hybrid_architecture}
\begin{figure}[tb]
	\centering
	\includegraphics[trim = 10 10 10 15, clip, width=\columnwidth]{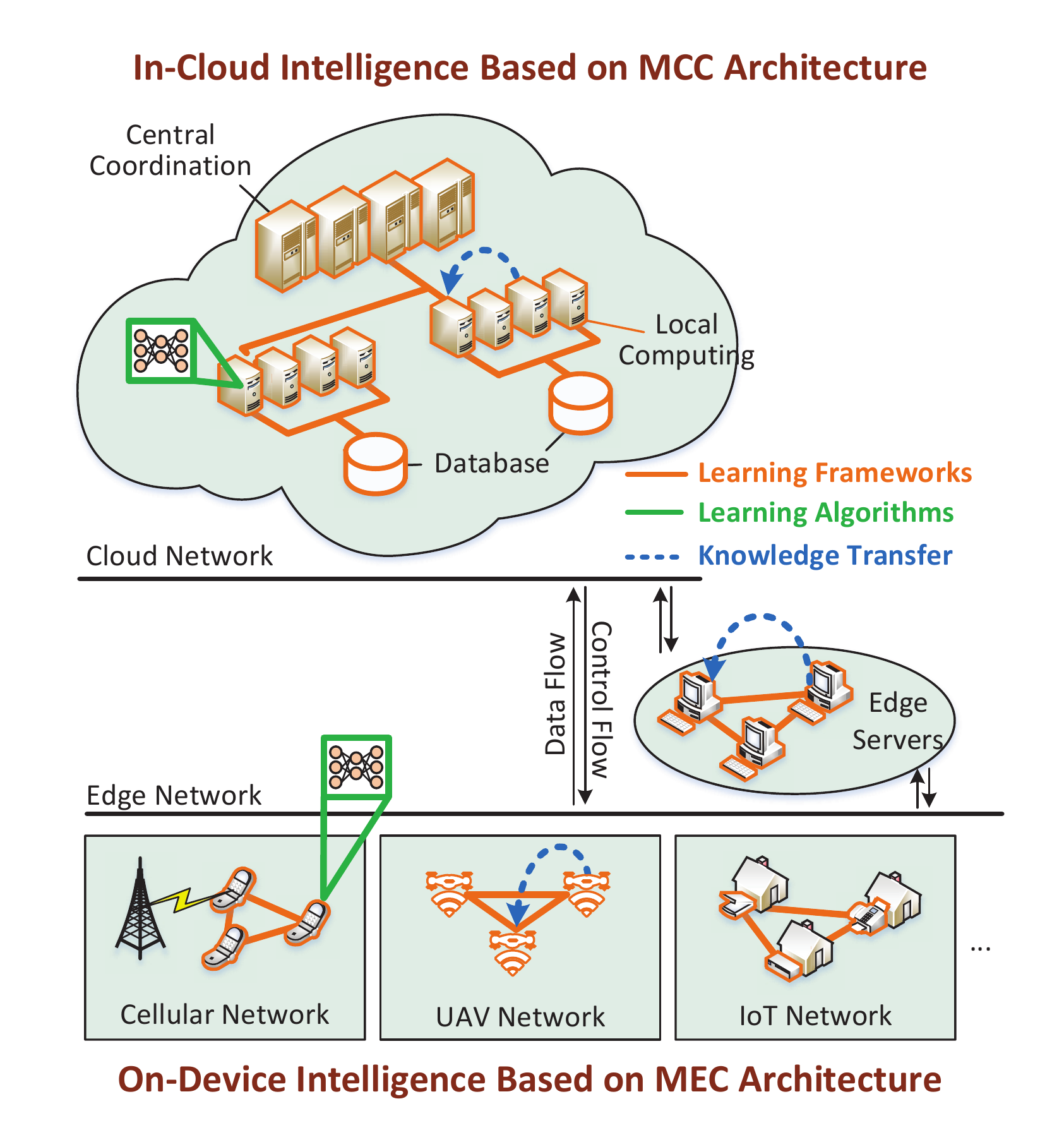}
	\caption{An integrated picture of the scalable data-driven wireless architecture based on MCC and MEC. The learning frameworks specify how distributed entities collaborate with each other; the learning algorithms specify how each local entity learns from its collected data; knowledge transfer improves the learning performance on individual nodes.}
	\label{fig:architecture}
\end{figure}

In this section, we investigate how to integrate the scalable intelligence into existing wireless architectures. In particular, the scalable intelligence orchestrates the \textit{in-cloud intelligence} and \textit{on-device intelligence} built upon the mobile cloud computing (MCC) architecture and the mobile edge computing (MEC) architecture, respectively, which facilitates accurate, timely, and end-to-end response to cutting-edge communication technologies such as infrastructure densification, mmWave
communication, and energy-efficient network management. This integrated scalable wireless architecture also offers the infrastructural foundation, e.g., communication, networking and computing, for the later discussed scalable learning frameworks and scalable learning algorithms. Figure~\ref{fig:architecture} illustrates a promising hybrid scalable wireless architecture to approach the blueprint of intelligence-everywhere in data-driven systems.

\subsection{In-Cloud Intelligence}
The in-cloud intelligence within the scalable architecture aims at overseeing all the learning and management tasks across the whole network and making decisions from a global view.
One tempting choice is to make use of the MCC architecture with the support from software defined networking~(SDN).
Specifically, MCC proposes to gather powerful computing resources at one place in order to bring convenience and efficiency due to joint management and coordination. 
In this way, the scalable architecture allows making full use of all available cloud computing resources to deeply analyze the network by learning from more data with a sophisticated model.
Moreover, the in-cloud environment supports on-demand computing resource provisioning and fast internal information exchange, which also lay a good foundation for scalable computations. 
Furthermore, the SDN can be integrated in a software-based cloud center to enable programmable and software-oriented configurations. In this way, the scalable architecture is able to re-configure the network smartly and timely according to the scalable learning demands.


\subsection{On-Device Intelligence}
The on-device intelligence within the scalable architecture aims at bringing intelligence closer to terminal devices with less or even no dependency on the remote cloud and therefore enjoys the following merits.
First, performing learning on local devices largely reduces the latency caused by interacting with a remote cloud, which is crucial for delay-sensitive applications requiring the ultra reliable low latency communication (URLLC). 
Second, keeping dataset securely stored on each device can largely relieve the privacy and security concerns by avoiding private information exposure.
Third, on-device learning alleviates the dependency on the connectivity to a remote cloud, which makes on-device intelligence more reliable to use for harsh scenarios. 
In this context, the emerging MEC techniques~\cite{8030322}, which promise to bring the computing and storage capability closer to the devices at the edge, can be exploited to construct the on-device intelligence in the scalable architecture.
Particularly, each edge device in the MEC architecture is able to acquire its own dataset through local sensing, and train its own learning model to tackle small-scale problems individually. 
As such, the scalable architecture can trade accuracy for latency (or response time) to local events on different devices.
Moreover, neighboring devices could collaborate with each other to contribute a superior collective intelligence through, for example, multi-agent learning.
In this way, the on-device intelligence is able to complement the in-cloud intelligence to offer end-users with prompt reactions, secured privacy, and seamless connectivity.

\section{Scalable Learning Frameworks} \label{sec:scalability}
\begin{figure}[tb]
	\centering
	\includegraphics[trim = 15 15 15 15, clip, width=0.9\columnwidth]{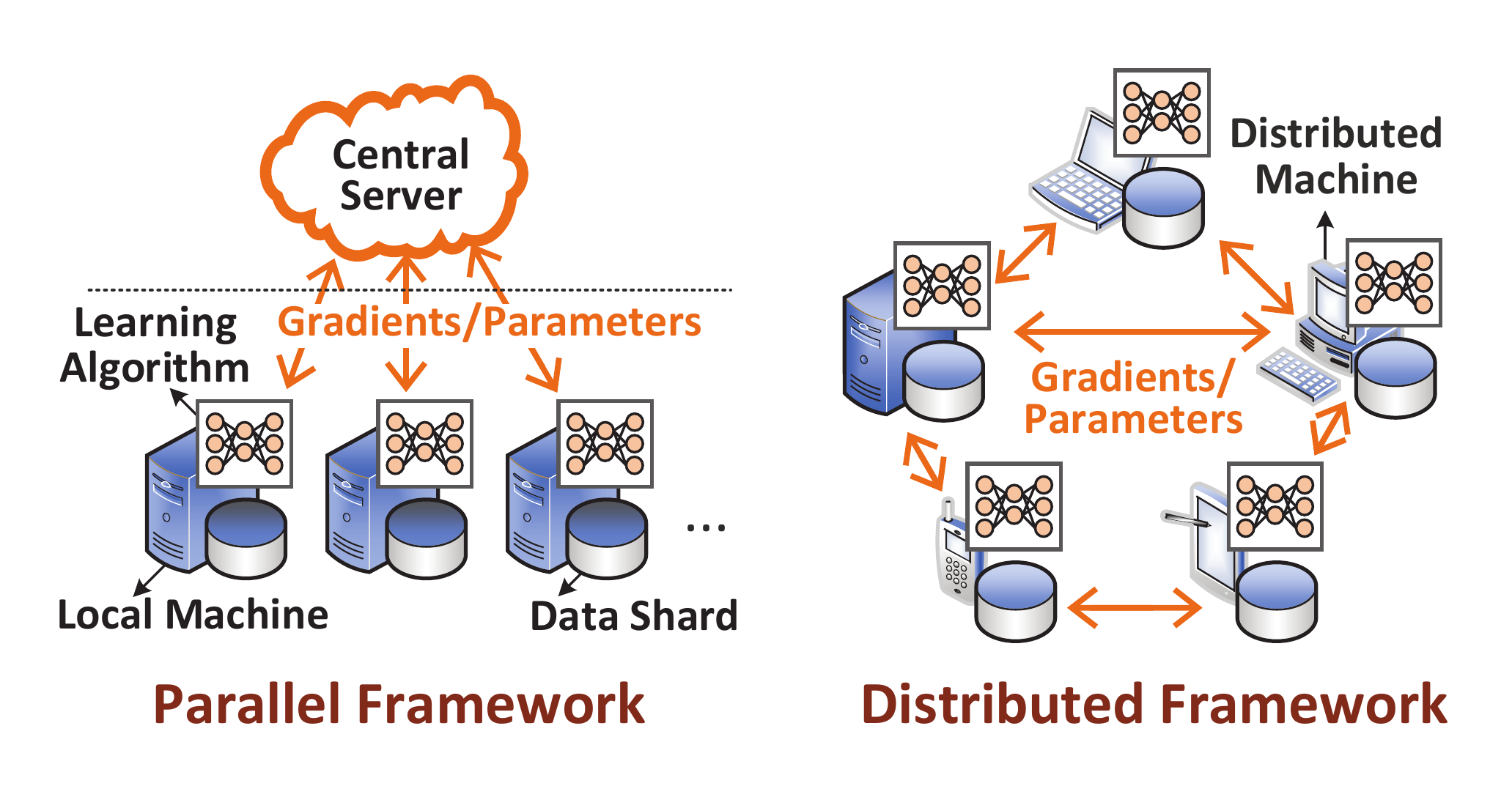}
	\caption{Computation topology of scalable learning frameworks, including a two-layer topology for parallel learning and a ring-shaped topology for fully distributed learning.}
	\label{fig:topology}
\end{figure}
Although the aforementioned scalable architecture demonstrates how to integrate scalable intelligence into the wireless infrastructure, it is unclear how distributed devices should learn jointly.
In this section, we present the scalable learning frameworks and specify the learning protocol for distributed devices. 
The scalable learning frameworks are classified into two categories according to their computing topologies, i.e., the parallel framework and the fully distributed framework.

\label{sec: scalable_framework}
\subsection{Parallel Learning Framework}
Parallel learning scales up the learning process by decoupling a large-scale learning problem into a bunch of small subproblems and handle them in a distributed manner. In this way, the computational burden can be distributed to multiple machines, thus improving the overall processing speed.
As shown in Fig.~\ref{fig:topology}, a representative parallel computation topology comprises two layers: (1) a bottom-layer with multiple local machines, where each of them learns from a subset of the complete dataset; (2) a top-layer with one (or several) global machine(s), which coordinates the learning at the bottom-layer and fuses the local results into a global one. 
As such, parallel frameworks require concurrent and iterative communications between the global and local computing machines due to frequent information exchange. In the sequel, we present two promising parallel computing frameworks, which are popular and applicable to machine learning algorithms.

\textit{1)~Alternating direction method of multipliers (ADMM)}
was first introduced in the mid-1970s and extended to handle a wide range of optimization problems largely in machine learning nowadays~\cite{Boyd11}. 
Specifically, ADMM takes the form of a decomposition-coordination procedure, which blends the benefits of dual decomposition and augmented Lagrangian methods for constrained optimization~\cite{Boyd11}. The basic idea behind the ADMM is to introduce a local copy of the global parameters for each local node and coordinate the local nodes to solve a large global problem through iterative local optimization and global consensus.
ADMM is well-acknowledged to be effective in terms of solving large-scale optimization problems~\cite{Boyd11, Hong16}. Hence, it is particularly promising to be used in the data-driven wireless system, in which distributed calibration/tuning of a huge, complicated network is highly demanded. 


\textit{2)~Parallel stochastic gradient descent (PSGD) methods} are extremely popular for its capability of solving large-scale deep learning problems.
It is widely used under the parameter server framework~\cite{Li2014} with the following workflow. Each worker (local machine) at the bottom layer computes the gradient of the model (short as gradient in the sequel) by using one mini-batch from the full dataset; the parameter server (central machine) at the top layer aggregates all the gradients to update a set of globally shared parameters; each worker synchronizes its local parameters to the global parameters periodically for consensus. 
PSGD methods admit the same convergence rate as the single-machine mini-batch stochastic gradient descent~(SGD)~\cite{Lian2017}, but at the cost of using a $ K $ times larger mini-batch. Moreover, they need to wait for the slowest learner to synchronize the update. 
Recent literatures have started to consider asynchronous PSGD methods, where the resulting parameter server can update global parameters without waiting for all learners to finish. 

In addition, federated learning as another applicable parallel learning framework has attracted great attention recently~\cite{konevcny2016federated}. Specifically, federated learning emphasizes strongly on mobile user's sole ownership, high security, and privacy-preservation of the data. In federated learning, data privacy can be achieved by the canonical secure multiparty computation or differential privacy, or by a more promising blockchain based encryption and decryption scheme. Therefore, federated learning is highly compatible with the MEC wireless architecture. Besides, when dealing with distributed data of the same structure, federated learning can similarly adopt the aforementioned ADMM or PSGD methods to train the algorithms.

\subsection{Fully Distributed Learning Framework}
The fully distributed framework is favored when the underlying network, e.g., a distributed wireless sensor network and an ad-hoc network, does not allow centralized control at all. As shown in Fig.~\ref{fig:topology}, its computation topology specifies no central machine but assumes that the distributed machines are connected via a (sometimes time-varying) communication graph. 
Therefore, designing a proper communication protocol to promote efficient information exchange is crucial for fully distributed learning.
Noticeably, the fully distributed variants of the two-layer learning frameworks such as decentralized ADMM and decentralized PSGD are under fast development.
Therefore, the data-driven wireless network should equip with both the parallel learning frameworks and their fully distributed variants in order to rapidly adjust the computing topology according to actual network topology or computing demands. This is extremely appealing when flexibility and adaptability are of great importance.
It is also noteworthy that the performance of the fully distributed variants sometimes can surpass their parallel versions~\cite{Lian2017}. This is an interesting observation and worth in-depth study in the future.

\section{Scalable Machine Learning Algorithms}
\label{sec:algorithms}
While the scalable learning frameworks specify how distributed nodes communicate and coordinate with each other from a global view, the machine learning algorithms discussed in this section demonstrate how to extract desired patterns from a given dataset on each node. 

Generally, the machine learning algorithms enable each node to predict future quantities, e.g., traffic variations and user movements, by discovering patterns from past data and using them for better planning, which is the key to building an autonomous and adaptive data-driven wireless system. The popularity of machine learning in the past five years is mainly due to the rise of deep learning centered around deep neural networks. However, the deep neural networks are difficult to interpret and unable to evaluate the modeling and prediction uncertainty explicitly, which largely hinders their application in the context of data-driven management. 
Therefore, one aim of this section is to draw public attention from the popular deep learning to Bayesian nonparametric learning and RL. Bayesian nonparametric learning is natural in representing uncertainty of prediction and decision made based on interpretable probabilistic models and smaller amount of data~\cite{Ghahramani15}. RL, on the other hand, embeds uncertainty measurement elegantly though maximizing an average long-term return~\cite{sutton2018reinforcement}. 

\subsection{Bayesian Nonparametric Algorithms}
%

\subsubsection{Model}
GP is defined to be a collection of random variables, any finite number of which follow a Gaussian distribution~\cite{RW06}. 
In particular, a GP model can be described by a mean function and a kernel function, and the latter determines the modeling power of the GP algorithm to a large extent. In order to make a kernel function full of expressive power and automatically adaptive to the data, universal kernels are highly demanded. Advanced universal kernels include (1) spectral kernels, (2) linear multiple kernels with automatic hyper-parameter search, and (3) deep kernels harnessing the neural network structure.
In contrast to existing deep learning models, the GP model provides a natural uncertainty region, which is critical to data-driven managements. Besides, GP models are robust to data overfitting even with small data set because they are marginalized over all possible parameters. This is a welcome feature for wireless systems as data query is often expensive to conduct, especially at the mobile user side.


Apart from the GP algorithm introduced above, the development of deep Bayesian neural networks is rapid in recent years~\cite{Ghahramani15}. In contrast to the conventional deep learning algorithms, the Bayesian counterparts are rather flexible as they do not need to fix the network structure, i.e., the number of hidden layers and the number of neurons, \textit{a priori}. By adopting nonparametric priors, these two parameters can be adjusted in light of the posterior distributions. The model hyper-parameters can be tuned by maximizing the evidence lower bound (ELBO) under the variational Bayesian setup. 

\subsubsection{Scalability}
The GP algorithms can be implemented in a distributed and principled manner, which makes it perfect fit to the data-driven wireless communication system. One could employ for each node a local machine, train a smaller scale GP model with a subset of the data, and merge the local kernel hyper-parameter estimates into a global one via consensus. Similarly, in the test phase, local predictions can be made and merged to a global one if necessary. In our recent work~\cite{8664622, Xie19}, we applied both the classical ADMM and a most recent proximal ADMM to speed up the hyper-parameter estimation process. The GP algorithms can be modified to adapt to online streaming data with rather low complexity as well.~

\subsubsection{Applications}
In the following, we give some representative wireless applications of non-parametric Bayesian algorithms, but our list is by no means complete. First, non-parametric Bayesian algorithms are particularly powerful for representing complicated mechanisms, such as 5G and V2X channel impulse responses, multi-path radio signal propagation, radio feature maps (such as signal quality, uplink/downlink traffic, and wireless resources demand/supply) evolving over time and space, and system error aggregation in a more economical and adaptive manner as compared with the state-of-the-art over-parameterized deep neural networks. Second, non-parametric Bayesian algorithms are more favorable to use in terms of system representation, control and integration. In particular, the GP algorithms can be combined with the traditional state space model to accurately represent and rebuild complicated trajectories generated by human beings, autonomous vehicles, and/or UAVs, and promising applications among others including very high-precision probabilistic fingerprinting and indoor/outdoor navigation based on wireless signals. Besides, Bayesian neural networks and deep GP algorithms can be combined with RL techniques to form Bayesian deep reinforcement learning (BDRL) to achieve a fully Bayesian control framework. Third, Bayesian nonparametric algorithms are also more natural to be used to fuse multi-modal data collected from different sensors, such as raw sensory data from smart phone motion sensors, ultra-sound from UAV ultrasonic sensors, and images and videos from surveillance camera, in different format and varying data qualities. Information dissemination and parameter inference can be carried out in a fully probabilistic manner. 

\subsection{Reinforcement Learning Algorithms}
\subsubsection{Model}
In RL, actions are taken to maximize the cumulative reward in an unknown environment, which is usually modeled as a Markov decision process~(MDP)~\cite{sutton2018reinforcement}. Different from supervised learning where the training data is usually labeled, an RL agent needs to decide what to do solely based on the environmental feedback, i.e., the reward signal, without labels.
In an RL model, the policy function defines the behavior of the RL agent by specifying the probability of taking an action at each state of the environment. The value function (or the Q-function) evaluates the total amount of rewards to accumulate over the future, conditioned on a starting state (or a starting state-action pair) and a policy to follow thereafter. 
The recent popular DRL adopts deep neural networks to approximate the policy function or the value function (or both) to greatly improve the generalization capability of RL when solving tasks with high-dimensional state and action spaces.
This makes DRL a promising tool for smart network management in the hybrid and complex data-driven wireless systems.

Generally, the DRL revolution starts from the emerge of deep Q-learning, which approximates the Q-function in Q-learning with deep neural networks, i.e., the Q-network. The success of deep Q-learning depends on two innovative training techniques: 
(1) train the Q-network with samples from a shuffled offline replay buffer to minimize sample correlations,
and (2) train the Q-network by following a target Q-network to give a consistent learning target. 
However, the deep Q-learning can only handle discrete and low-dimensional action spaces. 
Hence, the combination of deep learning and the actor-critic RL algorithm is later proposed to adapt DRL to the continuous action domain.
The representatives of this research line include 
(1)~the deep deterministic policy gradient (DDPG) algorithm, which replaces the conventional stochastic policy gradient with deterministic policy gradient to improve the learning efficiently,
(2)~the asynchronous advantage actor-critic (A3C) algorithm, which runs multiple DRL instances in parallel to perform policy updates without using an offline replay buffer,
and (3)~the proximal policy optimization (PPO) algorithm, which clips the policy changes per training step to improve the learning robustness.

\subsubsection{Scalability}
The DRL algorithms can be trained in parallel using the aforementioned parallel learning framework based on PSGD, and thus they perfectly  fit to the data-driven wireless system. 
One could employ for each distributed node an actor-learner pair to perform parallel RL training. The actor interacts with the environment and stores experiences in a replay buffer, and the learner samples experiences from the replay buffer and computes gradients with respect to model parameters. The computed gradients from each distributed node are then collected to update a set of global parameters. Finally, the local parameters are synchronized with the global parameters periodically for consensus. As an example, our recent work~\cite{xu2019load} trains the DRL model in parallel 
by employing multiple actor-learner pairs with multiple exploring strategies to learn the optimal policy for load balancing. Besides, the A3C and PPO algorithm can also be trained in parallel, without using a replay buffer.

\subsubsection{Applications} 
The representative directions for applications of DRL in data-driven wireless systems include network operation/maintenance, resource management and security enhancement. The context of MDP (e.g., state, action, and reward) should be properly specified in different applications.
In particular, for network operation/maintenance such as handover management and user localization, the DRL algorithms can be employed to decide the optimal actions of different operation/maintenance operations, e.g., handover and admission control. 
In contrast to traditional algorithms which are developed under prior assumptions on the system, e.g., user mobility patterns and network topologies, the DRL algorithms do not require any prior knowledge about the underlying environment thereby having a better capability to optimally configure the network under complex conditions.
For network resource management such as beamforming control, channel assignment, and network caching, the RL actions can be specified as  resource allocation operations, e.g., increasing or decreasing the transmitting power; the RL reward can be specified according to the performance metrics or the resource constraints, e.g., network throughput, communication latency, and quality-of-service (QoS). 
For network security enhancement, the DRL agent can be trained to recognize and avoid the network attacking attempts autonomously. For example, in jamming attacks, the attackers aim at sending jamming signals with high power to cause interference to the receivers. In this case, the DRL algorithms can be employed to estimate the attacker's jamming policy and respond adaptively.






.

\section{Data, Experience, or Both?}\label{sec:transfer}
In this section, we discuss how to improve the learning performance at each individual node through knowledge transfer.
First of all, in the field of wireless communication, traditional models based on mathematical modeling usually take advantage of valuable expert/domain knowledge on the target problem. Even if the traditional models may be not applicable to the target problem due to inaccurate modeling, their formulations can still help data-driven models to define a proper learning context thereby making complex tasks learnable, e.g., reward engineering in RL. In fact, formulating a good learning problem is as important as developing a good learning algorithm. Besides, traditional models can serve as the teacher of the data-driven models to improve the learning efficiency. For example, the learning agent can imitate the behaviors of the traditional models at the early learning stage to quickly reach a comparable performance and then continue to find superior alternatives. The knowledge transfer between traditional models and data-driven models in the context of wireless communication is still under exploration~\cite{zappone2019wireless}, which remains a challenging but promising research direction.

On the other hand, in the context of scalable learning, experienced learners and naive learners usually optimize a similar data-driven model, such that one could transfer knowledge from experienced learners to naive learners to improve the performance of scalable learning.
For example, the recent popular graph neural networks~\cite{zambaldi2018deep} can explicitly represent the structural information between task-relevant entities. In this case, task-correlated and task-specific knowledge can be modeled by different neural network layers, such that learners can reuse the task-correlated layers to reduce the training efforts.
Moreover, knowledge transfer among data-driven learners can also enhance system robustness, e.g., transferring knowledge from damaged devices to backup devices, and improve the data efficiency, e.g., transferring knowledge from existing devices to newly added devices. 

\section{Case Studies}
In this section, we provide a quantitative demonstration on how to use the presented scalable learning techniques to perform data-driven wireless communication and networking. Two cases are studied: (1) scalable GP-based wireless traffic prediction, and (2) scalable DRL-based load balancing.

\subsection{Case I: Scalable GP-based Wireless Traffic Prediction}
\begin{figure}[tb]
	\subfigure[Prediction accuracy]
	{ 	
		\label{fig:rmse}
		\includegraphics[trim = 5 5 5 5, clip, height=3cm]{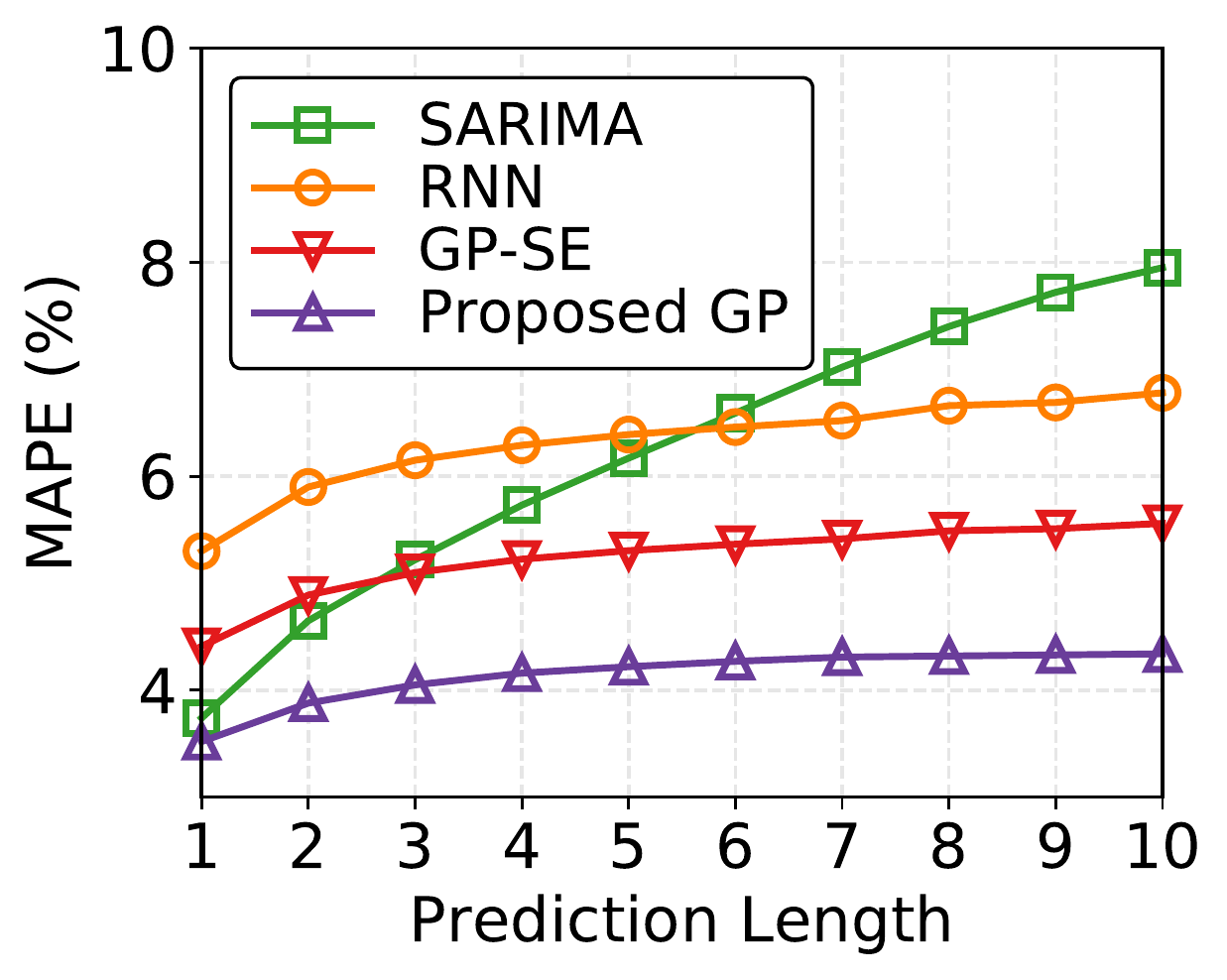}
	}
	\subfigure[Scalability]
	{ 	
		\label{fig:scalability}
		\includegraphics[trim = 5 5 5 5, clip, height=3cm]{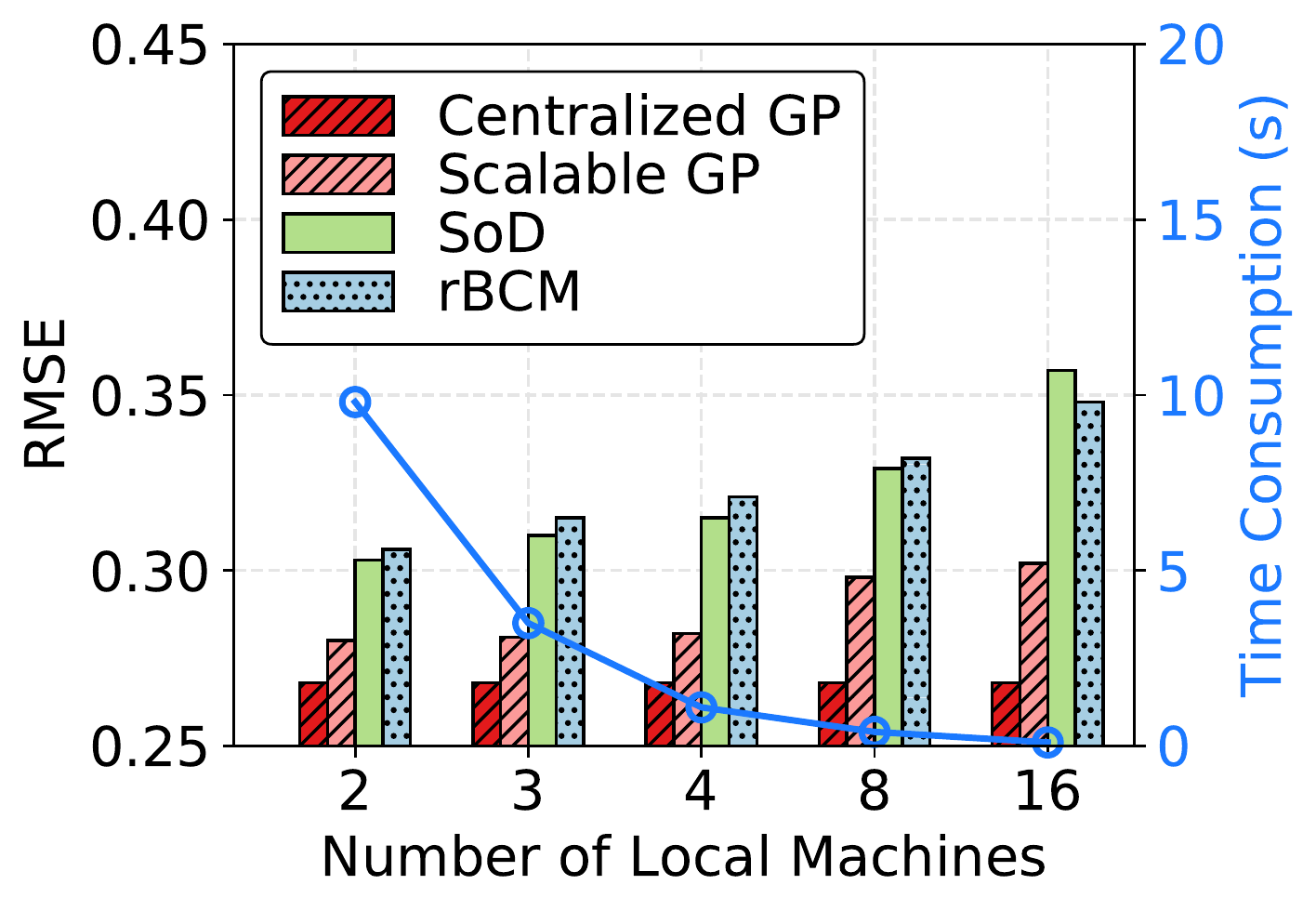}
	}
	\caption{Performance of the GP-based wireless traffic prediction. The experiments use real 4G datasets collected from one southern city in China.  (a) GP-SE uses only one SE kernel, while the proposed GP uses a composite kernel designed with domain knowledge. (b) Centralized GP trains the GP model using one machine with full dataset. The blue line describes the training time of the scalable GP.}
	\label{fig:case1}
\end{figure}
Wireless traffic prediction can effectively reduce the uncertainty in network demand and supply, and thus is a key enabler for smart management in data-driven wireless networks, e.g., traffic-aware base station on/off control~\cite{8304392}. Our recent work~\cite{8664622} presents a scalable GP-based traffic prediction framework based on the aforementioned MCC architecture and the ADMM framework to support large-scale traffic predictions in data-driven systems.
We choose a weekly periodic kernel, a daily periodic kernel, and a squared exponential (SE) kernel to model the weekly periodic pattern, the daily periodic pattern, and the dynamic deviations observed in real 4G traffics, respectively. As shown in Fig.~\ref{fig:rmse}, transferring such domain/expert knowledge into our kernel design can largely reduce the prediction error to as low as $ 3\% $ in terms of the mean absolute percentage error (MAPE), which outperforms other prediction algorithms such as seasonal autoregressive integrated moving average (ARIMA) and the deep recurrent neural network.
This verifies that the GP algorithms have a great potential to be widely applied in data-driven wireless systems.

Moreover, we proposed a scalable GP framework based on the classic ADMM framework to reduce the training complexity of standard GP from $ \mathcal{O}(N^3) $ to $\mathcal{O}(\frac{N^3}{K^3})$, where $ N $ is the number of training samples. Fig.~\ref{fig:scalability} shows 
the training time for the GP model using $ 700 $ traffic data points, and the training time can be reduced from $ 16.8 $ s to $ 0.1 $ s by increasing the number of distributed computing units from $ 2 $ to $ 16 $. 
Meanwhile, the prediction performance of our proposed scalable GP only degrades modestly compared with the standard GP with centralized training. 
We compare the prediction performance with two state-of-the-art low-complex GP models, namely the robust Bayesian committee machine (rBCM) which trains the model in a centralized manner but aggregates the prediction results distributively, and the subset-of-data (SOD) model which trains the model with one subset from the full dataset. The results in Fig.~\ref{fig:scalability} show that the scalable GP outperformed all other schemes, which verifies the superiority of our proposed scalable learning frameworks.
It is noteworthy that the proposed framework can also be turned into a fully distributed scheme based on MEC by allowing each computing unit to train its local GP model using a data subset individually~\cite{8664622}. 

\subsection{Case II: Scalable DRL-based Load Balancing}
Load balancing aims at automatically resolving the mismatch between network resource distribution and network traffic demand. It is becoming increasingly important in improving the resource utilization efficiency of data-driven wireless networks. Our recent work~\cite{xu2019load} presents a scalable DRL-based load balancing framework based on the aforementioned MCC architecture and the PSGD framework to handle the large-scale load balancing problem in a scalable manner. 
The proposed framework dynamically groups the underlying cells into different clusters and perform in-cluster load balancing with asynchronous parallel DRL. Each learning agent can autonomously accommodate its load balancing policy to irregular network topologies and diversified user mobility patterns. 
In particular, we define the MDP of the DRL-based load balancing model as follows: the state includes the information of cell load distribution and user distribution, the action is a handover parameter which controls the user handover among adjacent cells, and the reward signal is the inverse of the maximum load of all the cells, i.e., balancing the load distribution by alleviating the worst case. 

\begin{figure}[tb]
	\centering
	\subfigure[Averaged maximum cell load]
	{ 	
		\label{fig:rewards}
		\includegraphics[trim = 5 5 5 5, clip, width=0.95\columnwidth]{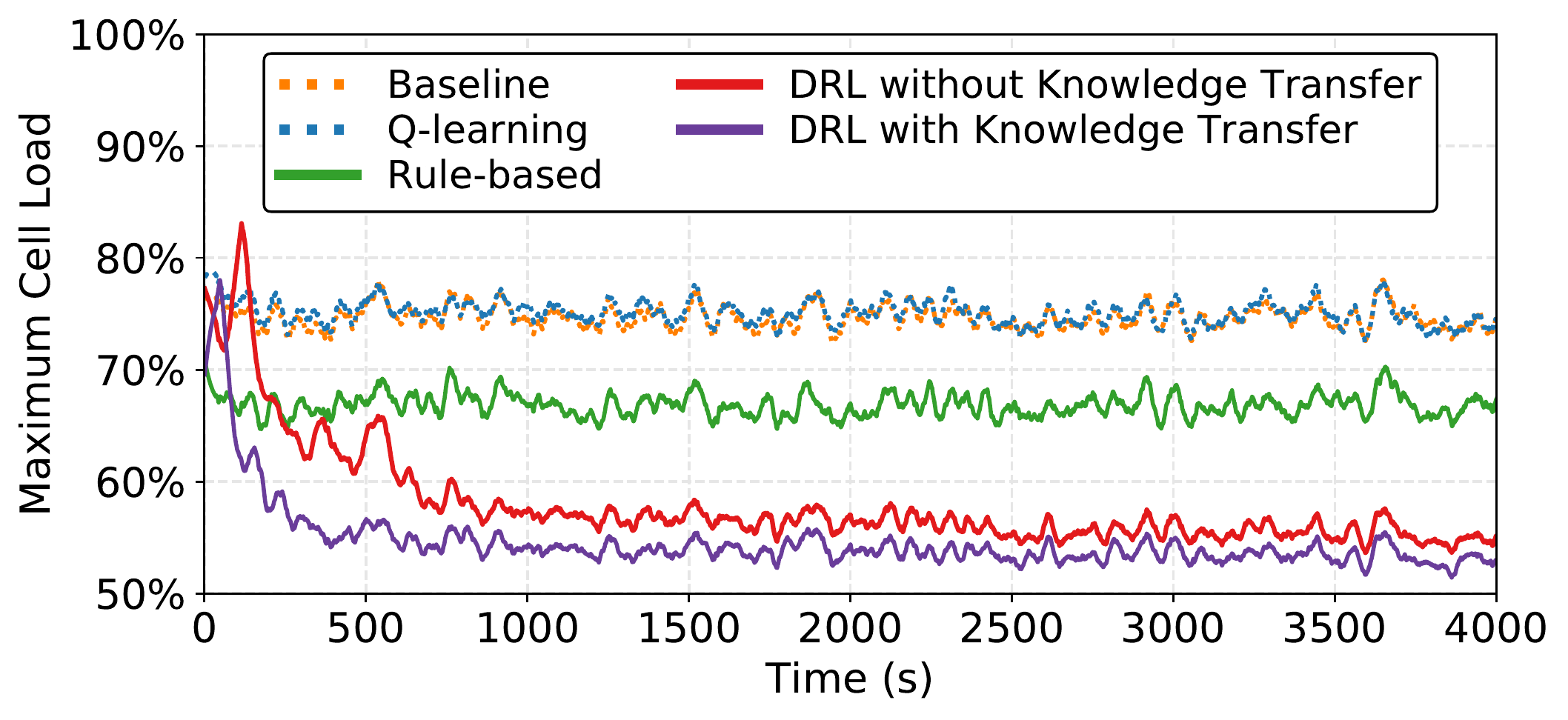}
	}
	\subfigure[Load standard deviation]
	{ 	
		\label{fig:std}
		\includegraphics[trim = 5 5 5 5, clip, width=0.88\columnwidth]{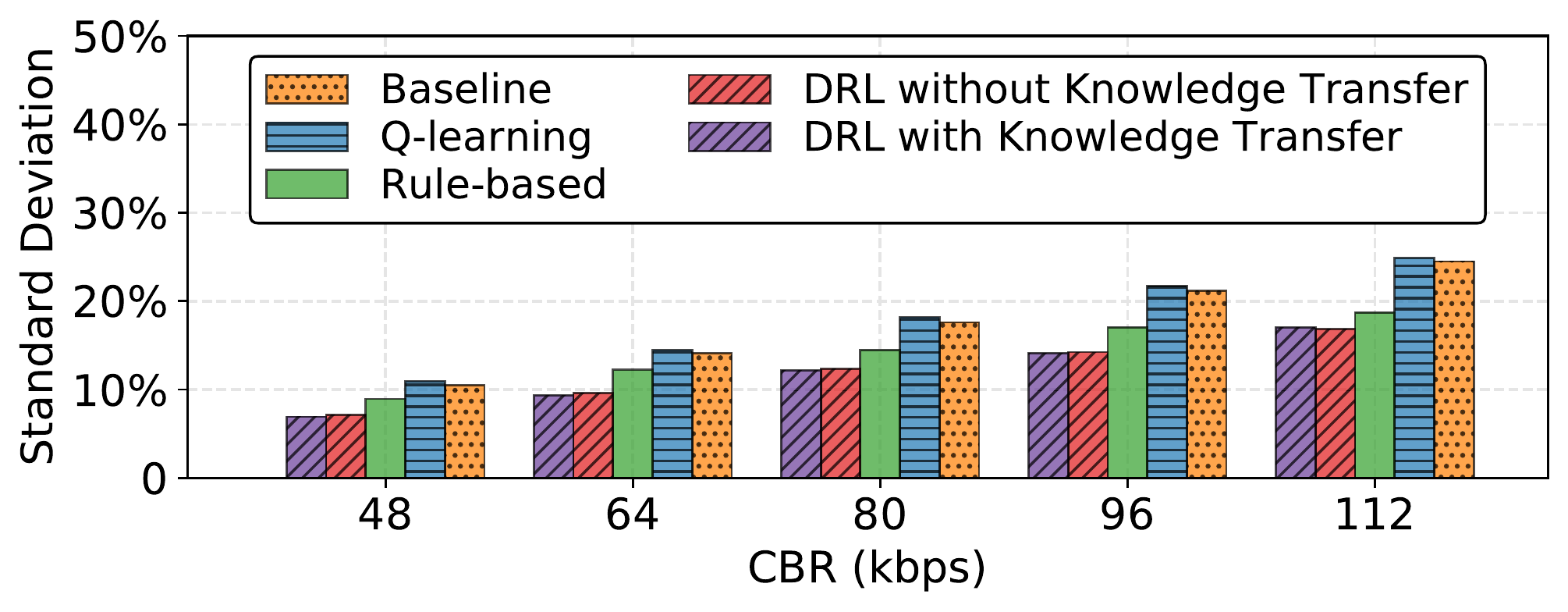}
	}
	\caption{Performance of the DRL-based load balancing. (a) The moving averaged maximum cell load over time steps. (b) The load standard deviations under different constant bit rate~(CBR) traffic demands. The simulated SON scenario consists of $ 12 $ small base stations randomly distributed in a $ 300~m \times 300~m$ area with $ 200 $ users randomly walking at $ 1~m/s$ to $10~m/s $, each incurring a constant bit rate traffic demand.}
	\label{fig:case2_result}
\end{figure}

Moreover, we propose to train the DRL algorithm under the aforementioned PSGD framework to improve the learning efficiency. Particularly, we employ multiple RL agents to interact with the wireless environment in parallel and allow them to share the gradients for joint learning, i.e., transferring knowledge across the agents. We also use traditional load balancing algorithms to generate high-quality training samples to guide the learning at early stage, i.e., transferring knowledge from traditional models. 
The results in Fig.~\ref{fig:case2_result} compare the performance of our proposed DRL-based load balancing model with other models, including the rule-based controller which balances the load by executing predefined rules, the Q-learning based controller which does not employ deep neural networks for generalization, and a plain baseline without performing any load balancing operations. The results show that (1) the proposed scalable DRL-based model substantially outperforms all other methods in terms of load balancing, and (2) transferring knowledge among the agents or from traditional models can improve the learning performance considerably.

\section{Discussions} \label{sec:future_challenges}
In spite of the apparent opportunities, there are challenges of applying scalable learning techniques to data-driven wireless networks.
First, we believe that in the future, computations will be distributed to edge devices to alleviate the load of the central node and reduce the latency. However, training an oversized model on such devices with limited computation power is impractical.
Therefore, it is necessary to develop low-complex learning models under the on-device constraints such as computing, storage and battery capabilities. 
Second, opening the black box of data-driven models to improve models' interpretability is surely a major topic for future investigation. Such interpretability can greatly help the researchers to (1) understand how data-driven models work, (2) design better scalable learning techniques, and (3) convince customers. 
\bibliographystyle{IEEEtran}
\bibliography{bib/IEEEabrv2015,bib/data_driven_mag_ref}

\begin{thebibliography}{10}
\providecommand{\url}[1]{#1}
\csname url@samestyle\endcsname
\providecommand{\newblock}{\relax}
\providecommand{\bibinfo}[2]{#2}
\providecommand{\BIBentrySTDinterwordspacing}{\spaceskip=0pt\relax}
\providecommand{\BIBentryALTinterwordstretchfactor}{4}
\providecommand{\BIBentryALTinterwordspacing}{\spaceskip=\fontdimen2\font plus
\BIBentryALTinterwordstretchfactor\fontdimen3\font minus
  \fontdimen4\font\relax}
\providecommand{\BIBforeignlanguage}[2]{{%
\expandafter\ifx\csname l@#1\endcsname\relax
\typeout{** WARNING: IEEEtran.bst: No hyphenation pattern has been}%
\typeout{** loaded for the language `#1'. Using the pattern for}%
\typeout{** the default language instead.}%
\else
\language=\csname l@#1\endcsname
\fi
#2}}
\providecommand{\BIBdecl}{\relax}
\BIBdecl

\bibitem{8304392}
W.~Xu, Y.~Xu, C.-H. Lee, Z.~Feng, P.~Zhang, and J.~Lin,
  ``Data-cognition-empowered intelligent wireless networks: Data, utilities,
  cognition brain, and architecture,'' \emph{{IEEE} Wireless Commun.}, vol.~25,
  no.~1, pp. 56--63, Feb. 2018.

\bibitem{8664622}
Y.~{Xu}, F.~{Yin}, W.~{Xu}, J.~{Lin}, and S.~{Cui}, ``Wireless traffic
  prediction with scalable {G}aussian process: Framework, algorithms, and
  verification,'' \emph{{IEEE} J. Sel. Areas Commun.}, vol.~37, no.~6, pp.
  1291--1306, Jun. 2019.

\bibitem{xu2019load}
Y.~{Xu}, W.~{Xu}, Z.~{Wang}, J.~{Lin}, and S.~{Cui}, ``Load balancing for
  ultra-dense networks: A deep reinforcement learning based approach,''
  \emph{{IEEE} Internet Things J.}, vol.~6, no.~6, pp. 9399--9412, Dec. 2019.

\bibitem{8030322}
N.~{Abbas}, Y.~{Zhang}, A.~{Taherkordi}, and T.~{Skeie}, ``Mobile edge
  computing: A survey,'' \emph{{IEEE} Internet Things J.}, vol.~5, no.~1, pp.
  450--465, Feb. 2018.

\bibitem{Boyd11}
S.~Boyd, N.~Parikh, E.~Chu, B.~Peleato, and J.~Eckstein, ``Distributed
  optimization and statistical learning via the alternating direction method of
  multipliers,'' \emph{Found. Trends Mach. Learn.}, vol.~3, no.~1, pp. 1--122,
  Jan. 2011.

\bibitem{Hong16}
M.~Hong, Z.~Q. Luo, and M.~Razaviyayn, ``Convergence analysis of alternating
  direction method of multipliers for a family of nonconvex problems,''
  \emph{SIAM J. Optim.}, vol.~26, no.~1, pp. 337--364, Jan. 2016.

\bibitem{Li2014}
M.~Li, D.~G. Andersen, J.~W. Park, A.~J. Smola, A.~Ahmed, V.~Josifovski,
  J.~Long, E.~J. Shekita, and B.-Y. Su, ``Scaling distributed machine learning
  with the parameter server,'' in \emph{Proc. 11th USENIX Symp. Oper. Syst.
  Des. Implement. (OSDI)}, Broomfield, CO, Oct. 2014, pp. 583--598.

\bibitem{Lian2017}
X.~Lian, C.~Zhang, H.~Zhang, C.-J. Hsieh, W.~Zhang, and J.~Liu, ``Can
  decentralized algorithms outperform centralized algorithms? {A} case study
  for decentralized parallel stochastic gradient descent,'' in \emph{Proc. Adv.
  Neural Inf. Process. Syst. (NIPS)}, Long Beach, California, USA, Dec. 2017,
  pp. 5336--5346.

\bibitem{konevcny2016federated}
\BIBentryALTinterwordspacing
J.~Kone{\v{c}}n{\`y}, H.~B. McMahan, D.~Ramage, and P.~Richt{\'a}rik,
  ``Federated optimization: Distributed machine learning for on-device
  intelligence,'' Oct. 2016. [Online]. Available:
  \url{https://arxiv.org/abs/1610.02527}
\BIBentrySTDinterwordspacing

\bibitem{Ghahramani15}
G.~Zoubin, ``Probabilistic machine learning and artificial intelligence,''
  \emph{Nature}, vol. 521, no.~1, pp. 452--459, May 2015.

\bibitem{sutton2018reinforcement}
R.~S. Sutton and A.~G. Barto, \emph{Reinforcement learning: An
  introduction}.\hskip 1em plus 0.5em minus 0.4em\relax Cambridge, MA, USA: MIT
  press, 2018.

\bibitem{RW06}
C.~E. Rasmussen and C.~I.~K. Williams, \emph{{G}aussian Processes for Machine
  Learning}.\hskip 1em plus 0.5em minus 0.4em\relax Cambridge, MA, USA: MIT
  Press, 2006.

\bibitem{Xie19}
A.~{Xie}, F.~{Yin}, Y.~{Xu}, B.~{Ai}, T.~{Chen}, and S.~{Cui}, ``Distributed
  {G}aussian processes hyperparameter optimization for big data using proximal
  {ADMM},'' \emph{{IEEE} Signal Process. Lett.}, vol.~26, no.~8, pp.
  1197--1201, Aug. 2019.

\bibitem{zappone2019wireless}
A.~{Zappone}, M.~{Di Renzo}, and M.~{Debbah}, ``Wireless networks design in the
  era of deep learning: Model-based, {AI}-based, or both?'' \emph{{IEEE} Trans.
  Commun.}, vol.~67, no.~10, pp. 7331--7376, Oct. 2019.

\bibitem{zambaldi2018deep}
V.~Zambaldi, D.~Raposo, A.~Santoro, V.~Bapst, Y.~Li, I.~Babuschkin, K.~Tuyls,
  D.~Reichert, T.~Lillicrap, E.~Lockhart \emph{et~al.}, ``Deep reinforcement
  learning with relational inductive biases,'' in \emph{International
  Conference on Learning Representations (ICLR)}, New Orleans, LA, USA, May
  2019, to appear.

\end{thebibliography}

\end{document}